\pdfoutput=1

\documentclass[11pt]{article}

\usepackage[]{acl}

\usepackage{times}
\usepackage{latexsym}

\usepackage[T1]{fontenc}

\usepackage[utf8]{inputenc}

\usepackage{microtype}
\usepackage{graphicx}
\usepackage{amsmath}
\usepackage{subcaption}

\title{Low-resource domain adaptation while minimizing energy\\and hardware resource consumption}

\author{
    Hernán Maina$^{1,2}$, Nicolás Wolovick$^{1}$, \and Luciana Benotti$^{1,2}$ \\
    $^1$FAMAF, Universidad Nacional de Córdoba, $^2$CONICET, Argentina\\
    \texttt{hernan.maina@mi.unc.edu.ar}\\
    \texttt{\{nwolovick,luciana.benotti\}@unc.edu.ar}\\
}

\begin{document}
\maketitle

\begin{abstract}
Training Large Language Models (LLMs) is costly in terms of energy, hardware, and annotated data, often resulting in a positionality rooted in predominant cultures and values~\cite{santy-etal-2023-nlpositionality}.
Domain adaptation has emerged as a promising strategy to better align models with diverse cultural and value contexts~\cite{hershcovich-etal-2022-challenges}, but its computational cost remains a significant barrier, particularly for research groups lacking access to large-scale infrastructure.
In this paper, we evaluate how the use of different numerical precision formats and data parallelization strategies impacts both training speed---as a proxy to energy and hardware consumption---and model accuracy, with the goal of facilitating domain adaptation in low-resource environments.
Our findings are relevant to any setting where energy efficiency, accessibility, or limited hardware availability are key concerns.
\end{abstract}

\section{Introduction}
In recent years, LLMs such as \textit{BERT}~\cite{devlin-etal-2019-bert} and \textit{T5}~\cite{JMLR:v21:20-074} have emerged as powerful tools across a wide range of Natural Language Processing (NLP) tasks. Their effectiveness largely stems from the \textit{fine-tuning} process, which enables the adaptation of \textit{pre-trained} models to specific downstream tasks using relatively small amounts of annotated data. Since these models acquire broad linguistic and statistical knowledge during pre-training, fine-tuning allows for efficient task specialization, often yielding strong performance.

However, when the domain of the target task differs substantially from the pre-training corpus, direct fine-tuning may not suffice. In such cases, a process known as \textit{domain adaptation}~\cite{daume-iii-2007-frustratingly}---also referred to as \textit{continued pre-training} or \textit{inter-training}---is typically employed. This method involves extending self-supervised pre-training by using a small amount of data from the new target corpus, enabling the model to better adapt its internal representations to the characteristics of the new domain. Prior work has shown both the benefits and trade-offs of this process when tailoring general-purpose models---like \textit{BERT}---to specialized contexts~\cite{ma-etal-2019-domain, gururangan-etal-2020-dont, peng-etal-2021-domain, grangier-iter-2022-trade}.

Domain adaptation of pre-trained language models facilitates personalization, but remains costly in terms of energy and hardware requirements. This poses a significant barrier for organizations without access to large-scale compute infrastructure---particularly in the Global South. Yet, the importance of computational efficiency transcends geography: optimizing resource usage also addresses environmental impact, institutional constraints, and global funding limitations.

\begin{figure}[!t]
  \centering
  \includegraphics[width=1\linewidth]{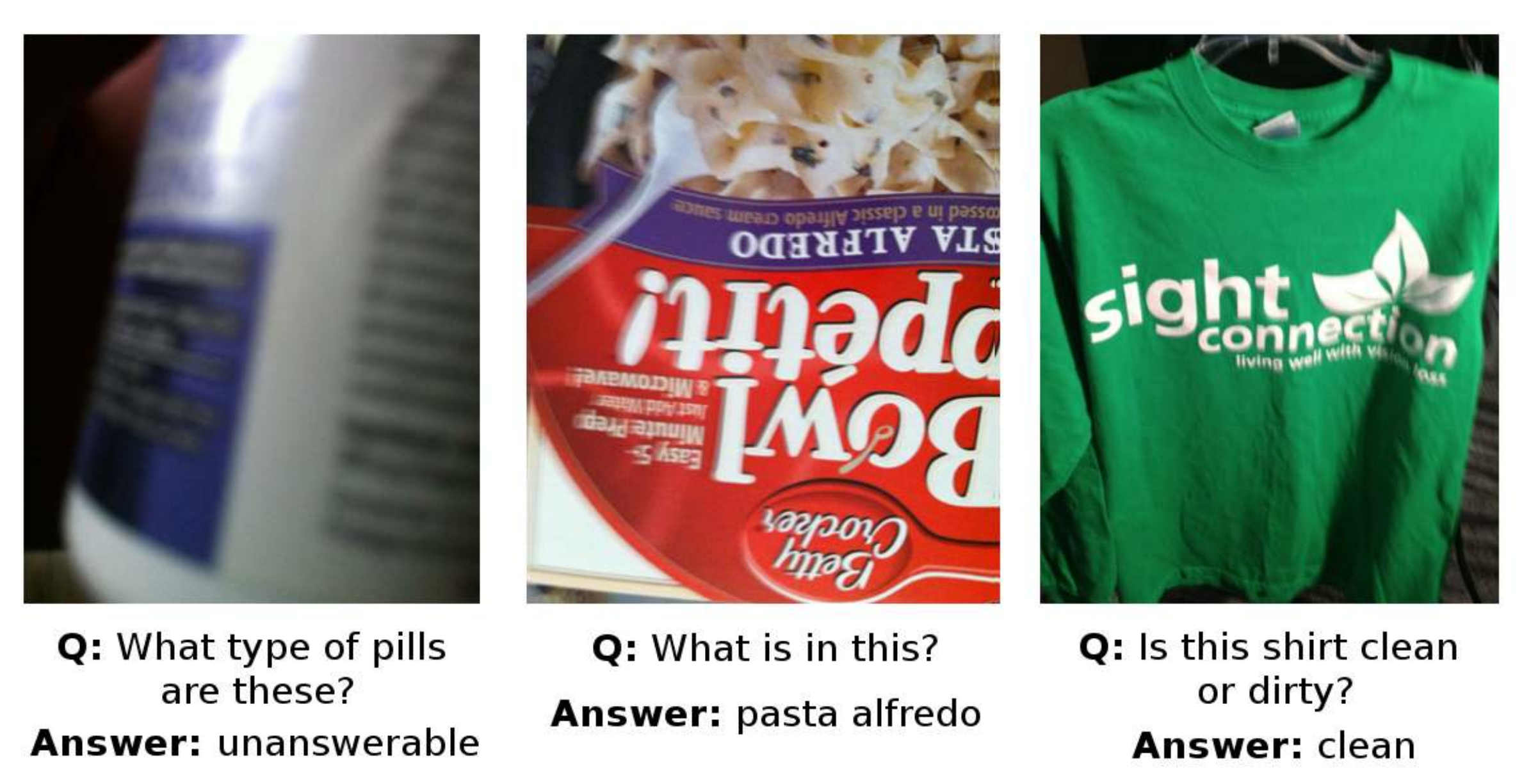}
    \caption{Examples of image pairs and associated visual questions retrieved from the VizWiz-VQA~\cite{DBLP:conf/cvpr/Gurari0SGLGLB18} dataset, illustrating challenges such as blurriness and poor lighting conditions leading to \textit{unanswerable} questions (left), unconventional framing and rotated text impairing answer visibility (center), and subjectivity in questions, such as assessing cleanliness (right).}
    \label{fig:vizwiz_examples}
\end{figure}

In this work, we explore how to make this adaptation process more efficient and accessible in low-resource settings. We focus on adapting the \textit{BERT$_{Base}$}\footnote{https://huggingface.co/google-bert/bert-base-uncased} model to the \textit{VizWiz-VQA}~\cite{DBLP:conf/cvpr/Gurari0SGLGLB18} dataset, which contains images paired with visual questions asked by people with visual impairments during their daily activities. Figure~\ref{fig:vizwiz_examples} presents representative samples from this dataset, highlighting key challenges such as blurring, rotated text, and subjective queries that can severely impact model performance.

Our goal is to improve downstream tasks such as Visual Question Answering (VQA), which holds significant assistive potential by supporting daily activities and promoting greater autonomy for people with visual impairments.

To pursue this goal, we focus on the linguistic backbone of VQA systems, using \textit{BERT} as a case study to explore the feasibility of efficient domain adaptation. Rather than exploring recent Parameter-Efficient Fine-Tuning (PEFT) techniques such as \textit{LoRA}~\cite{DBLP:conf/iclr/HuSWALWWC22} or \textit{QLoRA}~\cite{DBLP:conf/nips/DettmersPHZ23}---which optimize only a small subset of parameters or insert adapter layers---our work concentrates on strategies that can be seamlessly integrated into standard full-model training pipelines, without requiring architectural changes or additional libraries.

While our main focus is on domain-specific adaptation to \textit{VizWiz-VQA}---which allows the model to align with the context of this population, for example, the conversational style typical of questions posed by visually impaired users in the analyzed dataset---we acknowledge that this is merely a complementary step. It does not address the deeper structural biases and cultural imbalances embedded during pre-training, particularly those rooted in Global North-centric corpora, which remain an unresolved challenge for future work.

Within this scope, we evaluate strategies to optimize this adaptation in multi-GPU environments, focusing on the impact of floating-point precision and data parallelism on training time and performance. Our results show that 16-bit mixed precision, combined with distributed data parallelism, achieves significant speed-ups without compromising model accuracy.
\section{Background}
\label{sec:background}
In this section, we provide the necessary context to understand the foundation of our work. We begin by introducing the \textit{VizWiz}\footnote{https://vizwiz.org} project, which supplies the real-world data underpinning our experiments, followed by a technical overview of the \textit{BERT} architecture and its domain adaptation through self-supervised learning. Additionally, we cover relevant implementation considerations, including numerical precision formats and parallelism strategies crucial for efficient large-scale model training.

\subsection{The VizWiz project}
\textit{VizWiz} is widely recognized as the first project to introduce datasets and challenges specifically designed to promote the development of assistive artificial intelligence technologies for people with visual impairments. Its primary goal is to foster inclusive research by providing real-world data that reflects the lived experiences and needs of this population.

In particular, the \textit{VizWiz-VQA} dataset is a visual question answering dataset consisting of images captured by visually impaired individuals during their daily activities, accompanied by spoken questions---later transcribed---and human-annotated answers. Unlike conventional VQA datasets, this introduces unique challenges, with images often exhibiting quality and technical issues such as blurriness, poor lighting, and unconventional framing, as well as more complex visual difficulties such as recognizing partially visible text in incorrect orientations, and handling subjective or unanswerable questions (see Figure~\ref{fig:vizwiz_examples}).

These characteristics make VizWiz-VQA an invaluable resource for developing Computer Vision (CV) and NLP models that are robust, context-aware, and truly inclusive.

\subsection{The BERT architecture}
\textit{BERT} (Bidirectional Encoder Representations from Transformers)~\cite{devlin-etal-2019-bert} is a pre-trained language model introduced by Google, built upon the \textit{Transformer} architecture~\cite{NIPS2017_3f5ee243}. Unlike earlier models that processed text unidirectionally---either left-to-right or right-to-left---\textit{BERT} employs a fully bidirectional training approach, allowing it to learn contextualized representations of each token by jointly attending to its left and right context within a sentence.

The architecture comprises a stack of layers known as \textit{encoders}, which serve as the model’s fundamental building blocks. Each encoder consists of a \textit{multi-head self-attention} mechanism followed by a feed-forward neural network (FFNN). The self-attention component enables the model to capture complex contextual dependencies between tokens, while the FFNN introduces non-linear transformations that enrich these contextual representations.

There are two main variants of the model: \textit{BERT\textsubscript{Base}} and \textit{BERT\textsubscript{Large}}. The \textit{Base} version contains 12 encoder layers and approximately 110 million parameters, whereas \textit{Large} includes 24 layers and around 340 million parameters.

\textit{BERT} adopts a subword tokenization strategy based on the \textit{WordPiece} algorithm, which segments words into smaller units---tokens---to handle rare or out-of-vocabulary words more effectively while maintaining a manageable vocabulary size. For input representation, a special $[CLS]$ token is prepended to the sequence, followed by the sentence tokens, and terminated with a $[SEP]$ token, which also serves to separate sentence pairs when required. The output embedding of the $[CLS]$ token is typically used for classification tasks, while $[SEP]$ plays a crucial role in tasks involving sentence boundaries or context separation.

\subsection{Self-supervised learning}
The widespread success of LLMs has been driven not only by the availability of large-scale web data and substantial computational resources, but also by the rise of self-supervised learning as a powerful training paradigm. A key insight underlying this approach is that expert-labeled data is not strictly required to train effective models. Instead, supervision signals can be automatically extracted from raw, unlabeled text.

Domain adaptation leverages this principle by continuing the original pre-training process on domain-specific data in a self-supervised manner. This allows the model to internalize statistical regularities and contextual patterns that are more representative of the target domain, thereby improving performance on downstream tasks. In the case of \textit{BERT}, domain adaptation typically involves additional pre-training on its two canonical self-supervised target tasks: \textit{Masked Language Modeling} (MLM) and \textit{Next Sentence Prediction} (NSP).

\paragraph{Masked Language Modeling (MLM).} In this task, random tokens in the input sequence are masked, and the model is trained to predict the original tokens based on the surrounding context. 
For instance, given the sentence \textit{``Please, $[MASK]$ this shirt''}, and within the domain of questions posed by people with visual impairments, the model may learn to favor domain-relevant completions such as \textit{``describe''} over more generic alternatives like \textit{``wear''} or \textit{``take''}. This illustrates the benefit of domain adaptation in aligning the model's lexical choices with the linguistic practices of the target population.

Typically, 15\% of the tokens in each input sequence are randomly selected and replaced with the special $[MASK]$ token. The model is then trained to recover these masked tokens using the surrounding unmasked context. This objective encourages the development of deep semantic representations, which generalize well to a wide range of downstream tasks, from token classification to broader language understanding.

\paragraph{Next Sentence Prediction (NSP).} While optional in some recent training setups, NSP was integral to \textit{BERT}'s original design. Its objective is to help the model capture inter-sentential relationships. During training, the model is presented with pairs of sentences and must predict whether the second sentence follows the first in the original corpus (positive example) or is randomly sampled (negative example). By modeling sentence-level coherence and discourse flow, NSP enhances the model’s ability to perform tasks that require reasoning over multiple sentences, such as Natural Language Inference (NLI) and Question Answering (QA).

\subsection{Numerical precision}
The rapid growth of LLMs has led to architectures with hundreds of billions of parameters. This increasing model size imposes substantial demands on both training and inference, especially for researchers and institutions relying on conventional hardware. As the number of parameters grows, so do memory requirements, computational overhead, and energy consumption, fostering a growing dependence on multi-GPU infrastructures and specialized accelerators.

One of the most effective strategies for mitigating these constraints is the use of reduced numerical precision formats. Since a model's memory footprint is determined not only by the number of parameters but also by how each is represented, choosing more compact data types can significantly reduce memory usage, training time, and bandwidth requirements. The two most commonly used formats are 32-bit floating-point (FP32) and 16-bit floating-point (FP16), each with distinct technical properties.

FP32 is the standard floating-point format in deep learning, representing numbers with 32 bits: 1 bit for the sign, 8 bits for the exponent, and 23 bits for the mantissa. This format offers high numerical precision and a wide dynamic range, making it well-suited for a broad variety of tasks, especially where numerical stability is critical.

FP16, also known as half-precision, reduces representation to 16 bits: 1 bit for the sign, 5 bits for the exponent, and 10 bits for the mantissa. This compact format lowers memory and bandwidth requirements and allows for faster arithmetic on hardware that supports 16-bit operations. However, the reduced exponent and mantissa ranges can make FP16 more susceptible to numerical underflow, overflow, or loss of precision in certain scenarios. To mitigate these limitations, training is often performed using Automatic Mixed Precision (AMP)~\cite{DBLP:conf/iclr/MicikeviciusNAD18}, which dynamically combines FP16 and FP32 operations based on sensitivity to precision.

In addition to these, intermediate formats such as BF16 (Brain Floating Point) and TF32 (TensorFloat-32) have gained popularity. BF16 preserves the 8-bit exponent range of FP32 but reduces mantissa precision to 7 bits, allowing faster training with less risk of underflow compared to FP16. It is widely supported on Google TPUs and recent NVIDIA GPUs. TF32, introduced by NVIDIA for its Ampere architecture, provides a middle ground between FP16 and FP32 by using the FP32 exponent with reduced mantissa precision, accelerating matrix multiplications while maintaining numerical robustness.

\subsection{Parallelism strategies}
Parallelism plays a crucial role in optimizing the training of deep learning models, especially when utilizing Graphics Processing Units (GPUs). GPUs are highly efficient at performing parallel operations, which significantly accelerates the computational demands of training LLMs. There are two main strategies for parallelizing training: \textit{model parallelism} and \textit{data parallelism}.

\paragraph{Model parallelism.} In this, the model is divided into several parts, and each part is assigned to a different GPU. This approach is useful when the model is too large to fit into the memory of a single GPU. Each GPU processes a portion of the model and performs its respective computations.

\paragraph{Data parallelism.} In this, the model is replicated across all GPUs, and the training data is distributed among them. Each GPU processes its portion of the data and computes gradients, which are then aggregated to update the model. There are two variations of data parallelism: in \textit{Data Parallel} (DP), one GPU acts as the main device responsible for synchronization and updating the model, while in \textit{Distributed Data Parallel} (DDP), each GPU works independently, processing its own data and updating the model autonomously without a main GPU, eliminating the need for synchronization~\cite{DBLP:journals/csur/Ben-NunH19}.
\section{Methods}
\label{sec:methods}
To assess the model's domain adaptation performance across multi-GPU setups, we designed experiments involving different parallelization strategies, numerical precision formats, and parameter tuning approaches.

\subsection{Experimental setup}
All experiments were conducted by evaluate training performance on two NVIDIA A30 GPUs \textit{(24 GiB HBM2), 64 GiB RAM, and CPUs dual Intel Xeon E5-2680v2}. The experiments were implemented using the \textit{PyTorch Lightning}, a lightweight wrapper for \textit{PyTorch}~\cite{DBLP:conf/nips/PaszkeGMLBCKLGA19} framework.

\subsection{Training process}
We performed domain adaptation of the \textit{BERT$_{Base}$} model on a total of 32.842 questions from the \textit{VizWiz-VQA} dataset, which were randomly divided into two subsets: 80\% for training and 20\% for validation.
Training was carried out over 5 epochs, testing different \textit{batch sizes} and \textit{learning rates} to assess their impact on performance. For practical purposes, domain adaptation was limited to the first self-supervised pre-training task---Masked Language Modeling (MLM).

Given the relatively small size of \textit{BERT$_{Base}$} and the architectural constraints that make it difficult to split the model evenly across multiple GPUs, we focused exclusively on data parallelism strategies. Specifically, we compared \textit{Data Parallel} (DP) and \textit{Distributed Data Parallel} (DDP) approaches, which are better suited to our model and experimental setup.

We evaluated two numerical precision formats: standard 32-bit floating-point (FP32) and 16-bit floating-point (FP16) using Automatic Mixed Precision (AMP). Since BF16 is not supported across all GPU architectures, and that TF32 is limited to specific NVIDIA hardware, we prioritized approaches that are widely compatible with diverse GPU platforms and training frameworks such as \textit{PyTorch} and \textit{TensorFlow}~\cite{DBLP:journals/corr/AbadiABBCCCDDDG16}, aligning with our goal of exploring strategies accessible to resource-constrained institutions.

During training, GPU memory consumption, utilization, and power usage were periodically logged to enable later analysis and comparison of efficiency.

\subsection{Evaluation metric}
To measure training performance, we used the number of \textit{``\textbf{e}pochs \textbf{p}rocessed \textbf{p}er \textbf{m}inute''} (eppm), defined as:

\[
eppm = \frac{{\text{total completed epochs}}}{{\text{training time in minutes}}}
\]

This metric provides a quantitative measure of training speed, allowing the efficiency of different configurations and strategies to be compared. A higher \textit{eppm} value indicates faster training and better hardware utilization.
\section{Experiments and results}
We conducted an evaluation of the optimization strategies applied during the domain adaptation process, focusing on the impact of batch size, parallelization strategies, and numerical precision formats. The analysis covers key metrics such as training performance—measured in $e$pochs $p$rocessed $p$er $m$inute (\textit{eppm})—alongside memory usage, power consumption, and model convergence. This assessment aims to clarify the trade-offs between efficiency gains and model performance across different training configurations.

\subsection{Batch size effect}
\begin{figure}[t]
  \centering
  \includegraphics[width=1.0\linewidth]{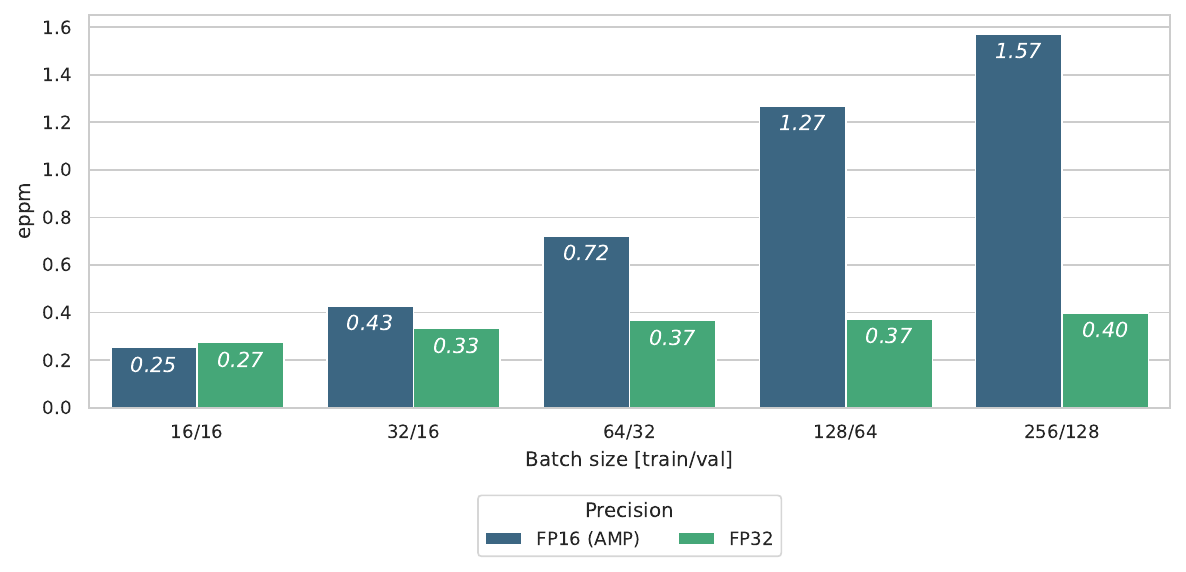}
    \caption{
    Impact of batch size on domain adaptation performance ($e$pochs $p$rocessed $p$er $m$inute) using a single GPU. Results are shown for both single precision (FP32) and mixed precision (AMP-FP16). 
    \textit{Performance improves consistently with larger batch sizes, with AMP-FP16 delivering up to 3.9× the throughput of FP32.}.
    }
    \label{fig:eppm_single_gpu}
\end{figure}

As a baseline, we evaluated the impact of batch size on training performance using a single GPU. As shown in Figure~\ref{fig:eppm_single_gpu}, increasing the batch size led to substantial performance gains, with AMP-based FP16 reaching up to 3.9× higher values than FP32 at the largest batch size tested.

\subsection{Parallelization strategies}
\begin{figure}[!t]
  \centering
  \begin{subfigure}[t]{\columnwidth}
    \centering
    \includegraphics[width=1.0\linewidth]{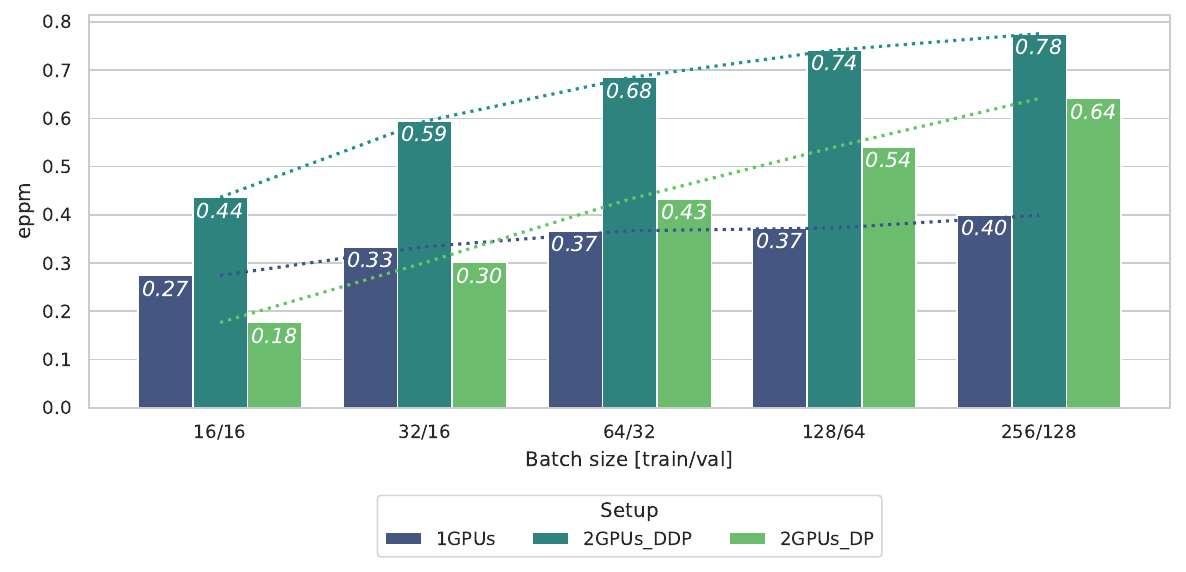}
    \caption{Performance comparison using FP32 precision.}
    \label{fig:eppm_fp32}
  \end{subfigure}

  \vspace{1em}

  \begin{subfigure}[b]{\columnwidth}
    \centering
    \includegraphics[width=1.0\linewidth]{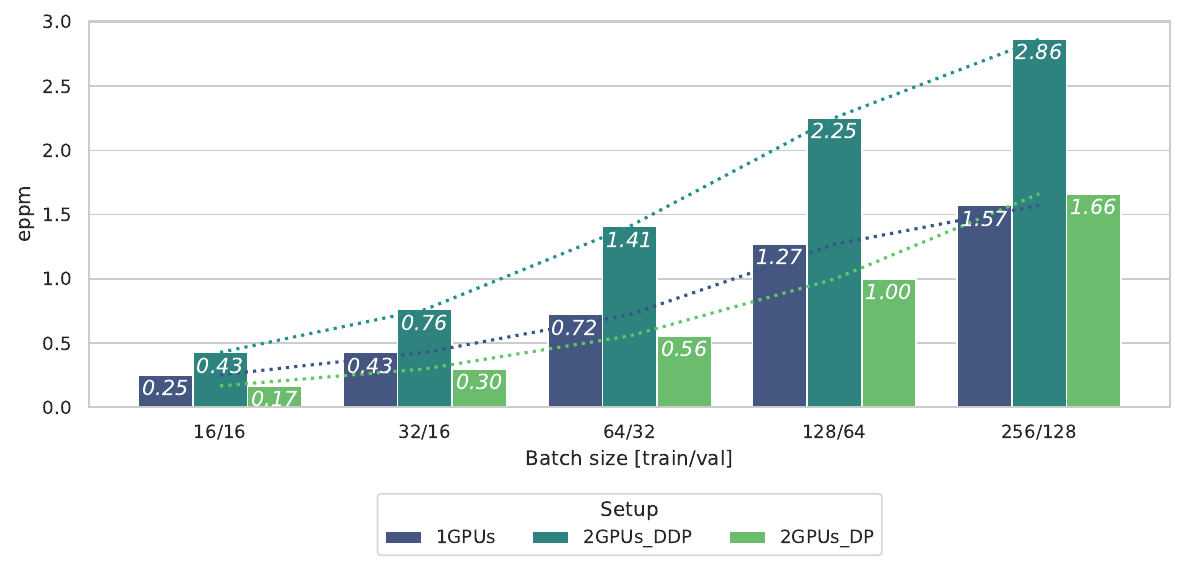}
    \caption{Performance comparison using AMP-FP16 precision.}
    \label{fig:eppm_fp16}
  \end{subfigure}
  
  \caption{
  Comparison of domain adaptation performance between FP32 (top) and AMP-FP16 (bottom) using single-GPU and multi-GPU setups with the \textit{Data Parallel} (DP) and \textit{Distributed Data Parallel} (DDP) strategies. Both plots show \textit{eppm} (epochs processed per minute) across different batch sizes. \textit{The results highlight the superior scalability of DDP, especially when combined with AMP-FP16, which consistently achieves higher throughput across all configurations.}
  }
  \label{fig:eppm_comparison}
\end{figure}

We compared baseline measurements obtained from domain adaptation using a single-GPU against the performance of multi-GPU setups with the DP and DDP strategies, under both FP32 and AMP-FP16 numerical precision formats. With FP32 (see Figure~\ref{fig:eppm_fp32}), DDP consistently outperformed DP across all tested batch sizes, achieving up to 2× the performance observed with a single GPU. Both parallelization strategies exhibited notable improvements over single-GPU training starting from a batch size of 64. However, for smaller batch sizes, synchronization and communication overhead between GPUs likely offset performance gains, making DP the least favorable option in these cases.

With AMP-FP16 (see Figure~\ref{fig:eppm_fp16}), DDP maintained its advantage, and its relative performance over single-GPU training improved further as the batch size increased. DDP showed up to 1.8× performance gains for the largest tested batch size, while DP generally underperformed compared to single-GPU setups across most batch sizes. This suggests that DP does not optimize inter-GPU communication as effectively as DDP, limiting its ability to fully exploit mixed-precision benefits.

Overall, both strategies (DP and DDP) achieved their best performance when using mixed precision instead of single precision. However, DDP combined with AMP-FP16 outperformed FP32 by up to 3.5×, compared to a maximum of 2.5× improvement achieved by DP+(AMP-FP16) over DP+FP32. This highlights the superior scalability and efficiency of DDP in multi-GPU environments, particularly when optimized with mixed precision.

\subsection{Memory usage analysis}
\begin{figure}[!t]
  \centering
  \begin{subfigure}[t]{\columnwidth}
    \centering
    \includegraphics[width=1.0\linewidth]{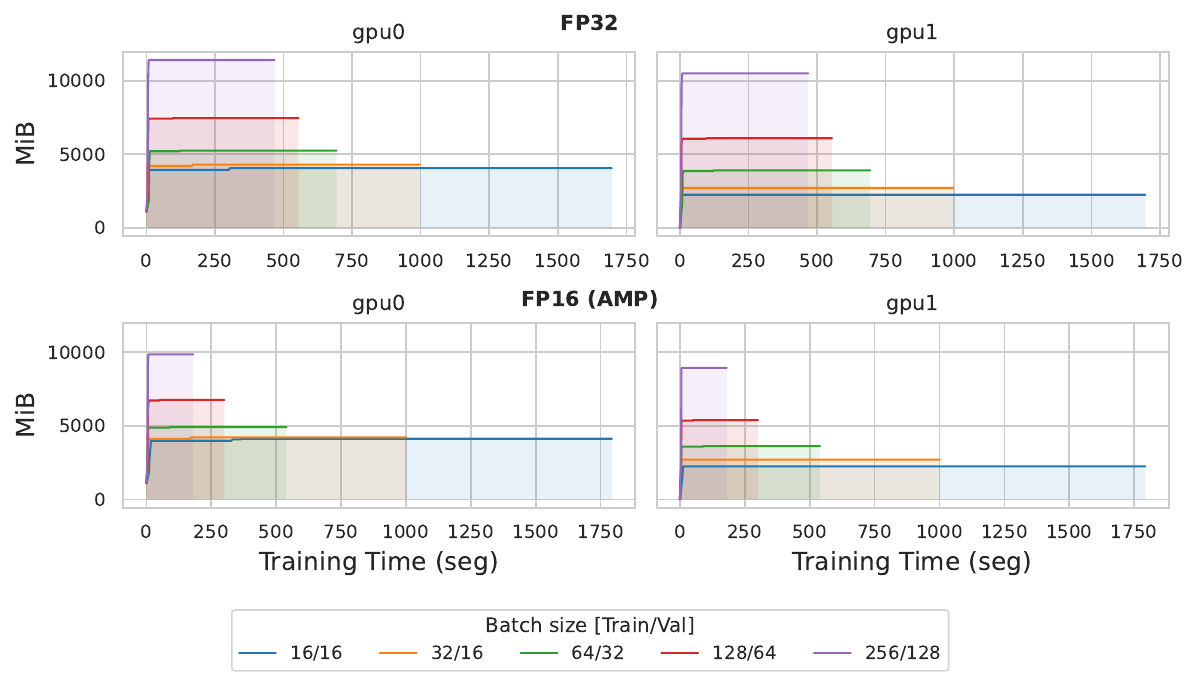}
    \caption{GPU memory distribution using DP.}
    \label{fig:dp_memory_usage}
  \end{subfigure}

  \vspace{1em}

  \begin{subfigure}[b]{\columnwidth}
    \centering
    \includegraphics[width=1.0\linewidth]{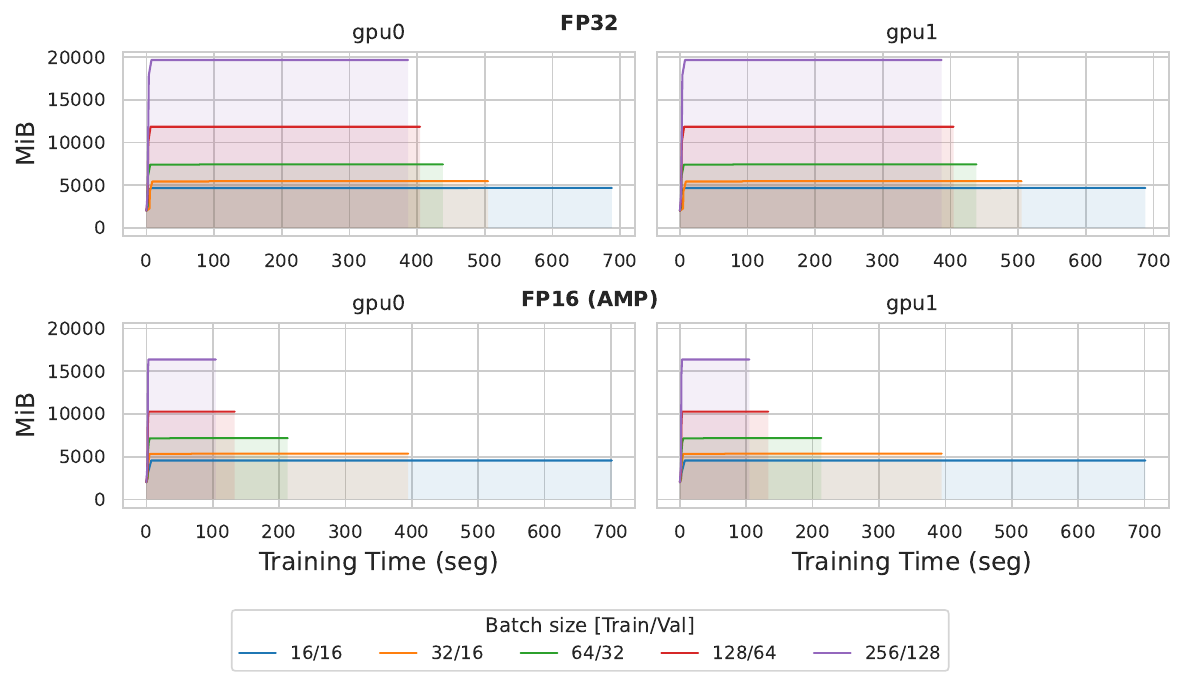}
    \caption{GPU memory distribution using DDP.}
    \label{fig:ddp_memory_usage}
  \end{subfigure}
  
  \caption{GPU memory usage comparison between \textit{Data Parallel} (DP, top) and \textit{Distributed Data Parallel} (DDP, bottom). Both plots compare memory consumption across different batch sizes and numerical precision formats (FP32 vs. AMP-FP16). \textit{This highlights a clear trade-off between memory usage and training speed}.}
  \label{fig:memory_usage_comparison}
\end{figure}

We periodically measured GPU memory usage during the domain adaptation process to evaluate how memory load is impacted by the DP and DDP strategies. Figures~\ref{fig:dp_memory_usage} and~\ref{fig:ddp_memory_usage} show the amount of memory (in MiB) used by each GPU throughout the adaptation process for the DP and DDP parallelization strategies, respectively.

The results reveal three distinct behaviors. First, both strategies (DP and DDP) exhibit a decrease in memory usage when using mixed precision compared to single precision. This difference becomes more pronounced as the training batch size increases. On average, a reduction of up to 15\% in memory usage was observed with DP and up to 20\% with DDP for both GPUs.

Second, a pattern related to the distribution of the memory load between GPUs is evident. In the DDP strategy, the memory load is distributed evenly across GPUs, whereas in the DP strategy, there is a clear imbalance, with \textit{gpu0} carrying a heavier load. This imbalance persists as the batch size increases. The reason behind \textit{gpu0} bearing more of the load can be attributed to the default configuration of the DP strategy, which designates one master GPU to coordinate data transfers and reductions during the training process.

Finally, we observe that DDP consistently uses more memory than DP across all configurations. This behavior is also a result of the default setup of DDP, which requires additional memory for processes involved in data distribution, synchronization, and gradient aggregation. However, as discussed earlier, DDP's ability to improve training speed through more efficient utilization of multiple GPUs leads to a trade-off between memory consumption and training velocity.

\subsection{Power usage and consumption}
\begin{figure}[!t]
  \centering
  \includegraphics[width=1.0\linewidth]{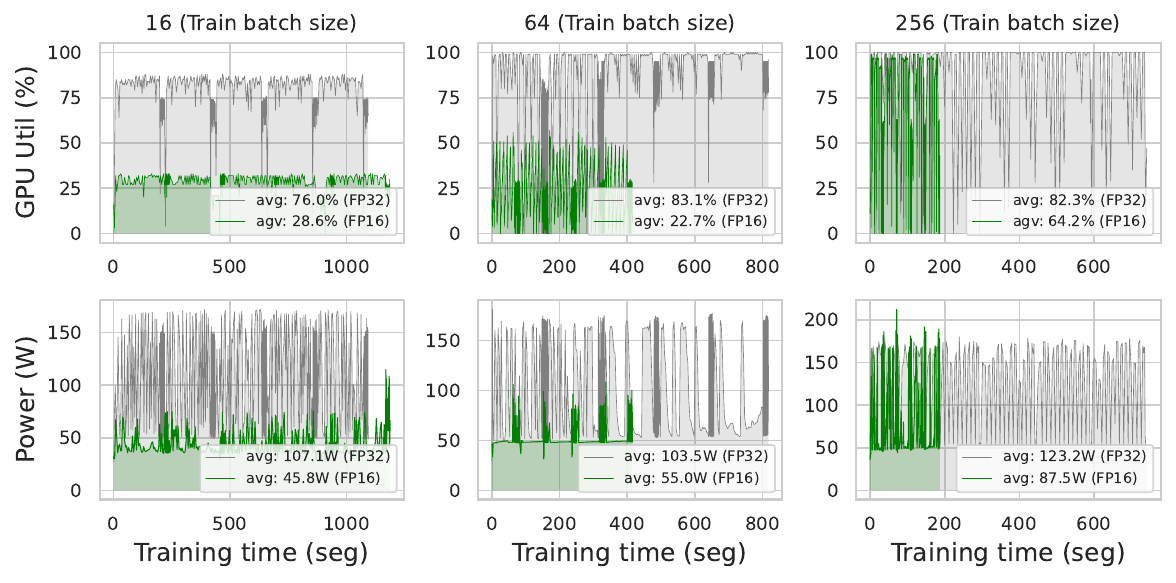}
    \caption{GPU utilization and power consumption during domain adaptation on a single GPU, comparing different batch sizes and numerical precision formats (FP32 vs. AMP-FP16). The top row shows GPU utilization over time, while the bottom row displays power consumption (in Watts). \textit{The results highlighting the reduction in both GPU load and energy consumption when using AMP-FP16.}} 
    \label{fig:baseline_power}
\end{figure}

As in the previous experiments, both power consumption (in Watts) and GPU utilization were periodically measured for each GPU involved. Figure~\ref{fig:baseline_power} presents the results for the domain adaptation scenario on a single GPU, showing how these values vary depending on batch size and numerical precision used. The figure displays, in its first row, the evolution of GPU utilization for batch sizes of 16, 64, and 256, while the second row illustrates the corresponding power consumption for those same sizes.

The results are consistent in both cases: both utilization and consumption decrease noticeably when AMP-FP16 is used instead of single precision, with a reduction of up to 2.7× in GPU utilization, which progressively diminishes as the batch size increases. This can be explained by the fact that small batches do not allow for efficient use of GPU resources, and when combined with lower computational load operations---as is the case with FP16---the underutilization of hardware becomes even more pronounced. Meanwhile, the bottom plots show that AMP-FP16 achieves an approximate 50\% average reduction in power consumption, regardless of batch size, suggesting that the lower computational demand associated with handling lower-precision data has a sustained energy impact.

\begin{figure}[!t]
  \centering
  \includegraphics[width=.95\linewidth]{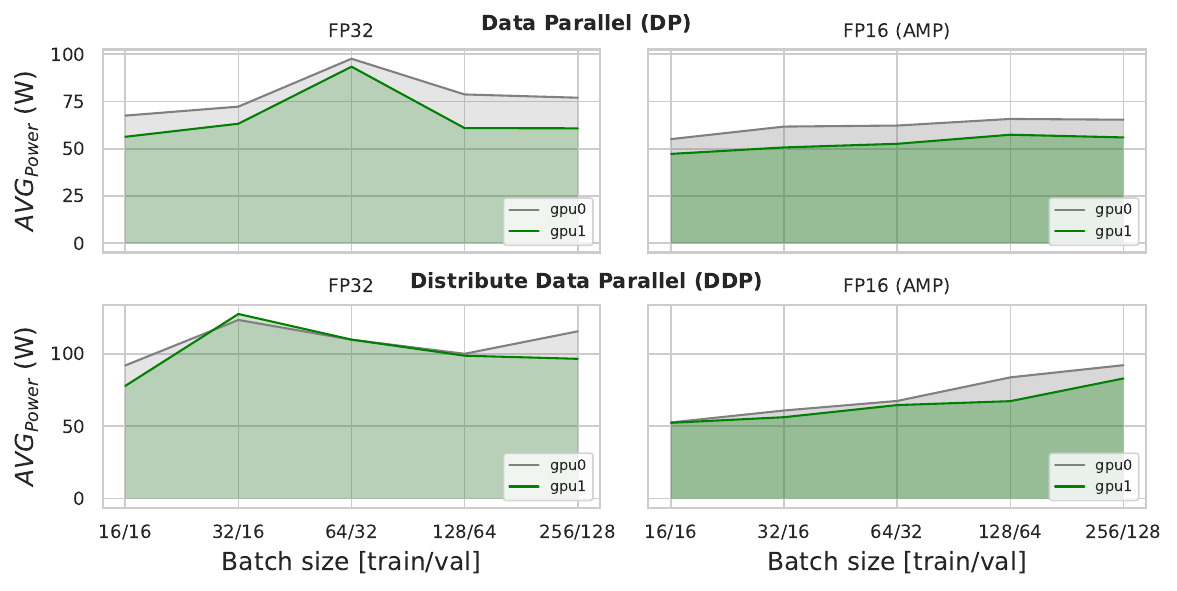}
    \caption{Power consumption (in Watts) during multi-GPU training for AMP-FP16 and FP32 across Data Parallel (DP) and Distributed Data Parallel (DDP) strategies. \textit{The results illustrate the energy savings achieved with AMP-FP16 and reveal the different consumption patterns between DP and DDP}.}
    \label{fig:dp_ddp_power}
\end{figure}

Figure~\ref{fig:dp_ddp_power} shows power consumption during training across multiple GPUs. In line with the previous results, the use of AMP-FP16 also reduces consumption in this scenario, replicating the trend observed with a single GPU. For the specific case of the DP strategy, as with memory utilization, an imbalance in the workload between GPUs is observed, with \textit{gpu0}---the master---bearing the bulk of the power consumption. In contrast, when using DDP strategy, no marked imbalance between GPUs is detected. However, the total power consumption is clearly higher than in DP, which can again be attributed to the additional computational overhead associated with creating independent processes and synchronizing gradients.

\subsection{Model convergence and accuracy}
\begin{figure}[!t]
  \centering
  \includegraphics[width=.95\linewidth]{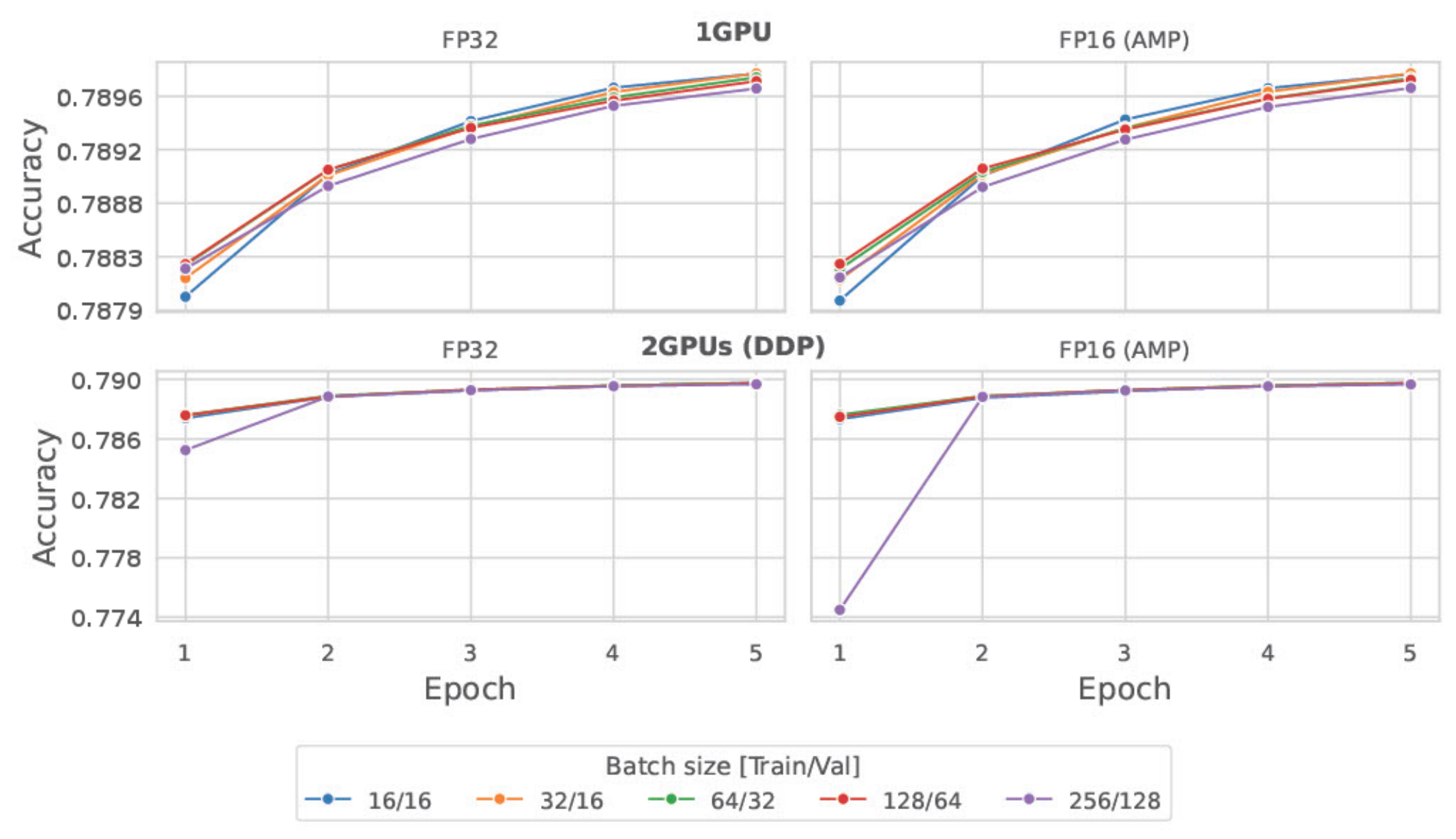}
    \caption{Validation accuracy curves across training epochs using the Distributed Data Parallel (DDP) strategy under both FP32 and AMP-FP16 numerical precision formats. The top plots correspond to single-GPU training, while the bottom plots show results with a dual-GPU setup. \textit{No significant differences in final accuracy were observed across the numerical precision formats or hardware configurations used}.}
    \label{fig:model_accuracy_convergence}
\end{figure}

Beyond efficiency metrics, we evaluated whether the proposed optimization strategies had any impact on model performance. To this end, we tracked masked word prediction accuracy on the validation set using the DDP strategy, which consistently delivered the best performance in previous tests.
We compared results between single-GPU and dual-GPU setups, evaluating both FP32 and AMP-FP16 precisions. All tests were conducted using the same batch size configurations as in earlier experiments, with learning rates properly scaled.
As shown in Figure~\ref{fig:model_accuracy_convergence}, no significant differences in final model accuracy were observed across the numerical precision formats or hardware configurations used. These findings suggest that using lower numerical precision is indeed feasible without compromising training quality.

\section{Conclusion}

In this study, we conduct a systematic evaluation of optimization strategies aimed at improving the efficiency of \textit{BERT} domain adaptation on GPU infrastructures. We compare \textit{Data Parallel} (DP) and \textit{Distributed Data Parallel} (DDP) strategies during the adaptation process across a range of batch sizes and numerical precision formats (FP32 and AMP-FP16).

Our results show that larger batch sizes consistently boosted performance, and that mixed precision (AMP-FP16) delivered up to 3.9× higher throughput in single-GPU settings, confirming its effectiveness as a lightweight acceleration technique. Among parallelization strategies, DDP proved the most scalable parallelization strategy, doubling single-GPU performance at best, and when combined with AMP-FP16, outperformed FP32 by up to 3.5×---outpacing DP’s improvements and reinforcing DDP’s superior efficiency.

We further analyzed GPU memory usage and utilization, finding that AMP-FP16 not only reduces memory consumption by up to 20\%, but also contributes to significantly more energy-efficient training. Specifically, we observed an average decrease of nearly 50\% in power draw (in watts) compared to FP32 in single-GPU settings, highlighting the dual advantage of AMP-FP16 in enhancing performance while reducing environmental impact.

Importantly, convergence evaluations indicated that lower numerical precision can be used without compromising the final quality of the domain adaptation, reinforcing the viability of mixed-precision training in practice.

Beyond technical performance, this study highlights the importance of considering the broader costs of model training, including environmental impact and accessibility, as emphasized by \citet{strubell-etal-2019-energy}. While our work is especially motivated by the computational constraints faced by research groups in the Global South, the proposed strategies are broadly applicable in any context where energy efficiency, sustainability, or limited hardware access are pressing concerns.
\section*{Limitations and future work}
All results presented in this study were obtained using a fixed hardware setup---two NVIDIA A30 GPUs---and a single model architecture (\textit{BERT\textsubscript{Base}}). As such, the trade-offs observed in performance, memory usage, and energy efficiency may differ when applied to other infrastructures, larger models, or different datasets. Extending the evaluation to a broader range of scenarios is a necessary step to assess the generalizability of our findings.

Our analysis focused on FP32 and AMP-FP16, as they are widely supported and commonly used in full-model training. While lower-bit formats like INT8 or INT4 offer appealing efficiency gains, they are typically optimized for inference or adapter-based fine-tuning, which fall outside the scope of this work. Future research could explore intermediate numerical precision formats such as BF16 and TF32, which promise further acceleration without compromising training stability.

Beyond precision formats, it would also be valuable to assess the impact of these optimization strategies on more recent language models---such as \textit{LLaMA}~\cite{touvron2023llamaopenefficientfoundation} or \textit{PaLM}~\cite{chowdhery2022palmscalinglanguagemodeling}---and in other domain-specific datasets. This would help validate whether the observed efficiency gains and trade-offs hold across different architectures and application contexts.
\section*{Ethical considerations}
Although this work focuses on exploring various optimization strategies for domain adaptation of the \textit{BERT} language model using the \textit{VizWiz-VQA} dataset---collected by visually impaired people---it is essential to recognize that even optimizations not directly aimed at improving model accuracy can subtly affect the stability and reproducibility of results. We are fully aware of the sensitivity involved in working with this population, and it is crucial to ensure that gains in efficiency do not come at the expense of model reliability, particularly because this population cannot easily verify the accuracy of the outputs. Therefore, any deployment of models trained under these optimizations should be preceded by a careful evaluation of the potential real-world consequences.
\section*{Acknowledgments}
We thank Julian Martin Eisenschlos for his comments and observations, which were valuable for refining this work.

This work used computational resources from UNC Supercómputo (CCAD) – Universidad Nacional de Córdoba~\footnote{https://supercomputo.unc.edu.ar}, which are part of SNCAD, República Argentina.

\bibliography{bib/custom}
\bibliographystyle{bib/acl_natbib}

\end{document}